\documentclass[a4paper,12pt]{article}

\usepackage{times}
\usepackage{mathptm}
\usepackage{amsmath,amsopn}
\usepackage{amsthm}
\usepackage{amsgen}
\usepackage{amssymb}
\usepackage{amstext}
\usepackage{url}
\usepackage{natbib}
\usepackage[T1]{fontenc} 

\usepackage[latin2]{inputenc}
\newcommand{\kwad}[1]{\left[\, #1 \,\right]} 
\newcommand{\klam}[1]{\left\{ #1 \right\}} 
\newcommand{\abs}[1]{\left| #1 \right|}

\newcommand{\oh}[1]{\, \overline{#1} \,}
\newcommand{\const}{\mathop{\text{const}}}
\newcommand{\junk}[1]{}
\newcommand{\boole}[1]{\left[\mkern-3mu\left[\,#1\,\right]\mkern-3mu\right]}

\DeclareMathOperator{\LZ}{LZ}
\DeclareMathOperator{\MDL}{MDL}
\DeclareMathOperator{\SEQUITUR}{SEQUITUR}
\DeclareMathOperator{\NSRPS}{NSRPS}
\DeclareMathOperator{\len}{len}
\newcommand{\length}[1]{\len #1}

\newtheoremstyle{plainenglish}{0.5ex}{0.5ex}{\it}{0pt}{\bf}{}{ }%
   {\thmname{#1 }\thmnumber{#2} \thmnote{(#3)}}
\theoremstyle{plainenglish}
\newtheorem{conjecture}{Conjecture}
\newtheorem{theorem}[conjecture]{Theorem}
\newenvironment*{proof*}[1]{\begin{trivlist}\item[]
\noindent\textbf{Proof of Theorem #1:}}{ \hfill\qedsymbol\par\end{trivlist}}

\author{{\L}ukasz D\k{e}bowski \\
  Institute of Computer Science \\
  Polish Academy of Sciences \\
  \url{ldebowsk@ipipan.waw.pl}}

\title{ On Hilberg's Law and Its Links \\ 
with Guiraud's Law }

\date{}

\begin{document}
\maketitle

\begin{abstract}
  \citet{Hilberg90} supposed that finite-order excess entropy of
  a~random human text is proportional to the square root of the text
  length.  Assuming that Hilberg's hypothesis is true, we derive
  Guiraud's law, which states that the number of word types in a~text
  is greater than proportional to the square root of the text length.
  Our derivation is based on some mathematical conjecture in coding
  theory and on several experiments suggesting that words can be defined
  approximately as the nonterminals of the shortest context-free
  grammar for the text.  Such operational definition of words can be
  applied even to texts deprived of spaces, which do not allow for
  Mandelbrot's ``intermittent silence'' explanation of Zipf's and
  Guiraud's laws.  In contrast to Mandelbrot's, our model assumes some
  probabilistic long-memory effects in human narration and might be
  capable of explaining Menzerath's law.
  \\
  \textbf{Keywords:} excess entropy, grammar-based compression,
  Guiraud's law, Zipf's law
  
\end{abstract}

%\tableofcontents

\section{Introduction}
\label{SecIntroduction}

Over a decade ago, \citet{Hilberg90} reinterpreted
\citeauthor{Shannon50}'s (\citeyear{Shannon50}) well-known experimental
data and formulated a novel hypothesis concerning the entropy of human
language.  The hypothesis states that block entropy $H(n)$ of a text
drawn from natural language production, except for disputable constant
and linear terms, is proportional to the square root of the text
length $n$ measured in phonemes (or letters),
\begin{align}
\label{HilbergH}
H(n)\approx h_0 + h_\mu n^{\mu} + h n,
\end{align}
where $\mu\approx 1/2$.  For brevity, we call relation
(\ref{HilbergH}) Hilberg's law.  Hilberg's publication appeared in
a~technical journal of telecommunications. It was popularized among
natural scientists by Ebeling \citep{EbelingNicolis91,EbelingPoschel94}
and stimulated some discussions
\citep{BialekNemenmanTishby00,CrutchfieldFeldman01,Shalizi01,Debowski01b,Debowski02}.

In this article, we shall discuss some interaction between Hilberg's
law and the better known Guiraud's and Zipf's laws.  Empirical
Guiraud's law \citep{Guiraud54} states that the number of orthographic
word types $V$ in a~text behaves like
\begin{align}
\label{Guiraud}
  V\propto N^\rho,
\end{align}
where $\rho<1$ is constant and $N$ is the length of the text
measured in orthographic word tokens. On the other hand,
Zipf's-Mandelbrot's law \citep{Zipf35,Zipf49,Mandelbrot54} states that
any text obeys relation
\begin{align}
\label{Zipf}
c(w)\propto \frac{1}{r(w)^{B}},
\end{align}
where $B>1$ is constant, frequency $c(w)$ is the count
of word $w$ in the text, and rank $r(w)$ is the position of word $w$
in the list of words sorted in descending order by $c(w)$.

We do not know to what extent Hilberg's law is valid. Formula
(\ref{HilbergH}) presupposes some stationary probabilistic model of
the entire natural language production, which is a~highly hypothetical
entity itself.  Nevertheless, we would like to argue that some form of
Guiraud's law can be deduced from equation (\ref{HilbergH}).  Strictly
speaking, assuming that Hilberg's law is true for all $n$, we shall
only infer some lower bound for the growth of the vocabulary size.
Despite that restriction, we think that our explanation of Guiraud's
law can be more linguistically plausible than the famous joint
derivation of Guiraud's and Zipf's laws provided by
\citet{Mandelbrot53}. The latter derivation is known also as
``intermittent silence'' explanation \citep{Miller57,Li98}.

Hilberg's law concerns the probabilistic distribution of arbitrary
phoneme or letter strings, i.e.\ the law constrains the distribution
of all human texts.  On the other hand, both Guiraud's and Zipf's laws
concern the distribution of individual words in texts.  Saying that
Guiraud's law can be deduced from Hilberg's law, we presuppose some
procedure which transforms the distribution of phoneme strings (i.e.\
texts) into the corresponding distribution of words. In some naive
approach, we could assume that the text is a~string of phonemes or
spaces and the words are the space-to-space strings of phonemes.  In
fact, ``intermittent silence'' explanation assumes that the text is
a~string of probabilistically independent random tokens taking the
values of spaces and phonemes.  Given this assumption and the
space-to-space definition of word, Mandelbrot deduced Zipf's law, and
hence Guiraud's law can be deduced as well \citep{Kornai02}.

Unfortunately, ``intermittent silence'' explanation cannot be applied
to natural language. We know that the occurrences of phonemes in the
language production exhibit some strong probabilistic dependence and
there are no definite spaces between the words in human speech
\citep{Jelinek97}. If we want to derive Zipf's law from the
distribution of mere phoneme strings, we must use some definition of
word tokens which could be applied to the text deprived of spaces and
which would match empirically the definition of word tokens given by
spelling conventions or by semantic considerations.

Some well-defined tokenization of the space-deprived text into
word-like strings can be given by grammar-based text compression
\citep{KiefferYang00}.  In grammar-based compression, the text is
represented as a~special context-free grammar, called an admissible
grammar. That class of context-free grammars should not be confused
with phrase structure grammars: The nonterminals of admissible
grammars correspond to fixed strings of phonemes rather than to
part-of-speech classes.  Each admissible grammar gives some
tokenization of the text into hierarchically structured word-like
strings being the nonterminal tokens. It was empirically confirmed
that for the grammars which approximate the shortest admissible
grammar for a~human text, the nonterminals usually correspond to the
orthographic words \citep{DeMarcken96,NevillManning96}.
 
We will show that the expected number of nonterminal types for the
shortest admissible grammar cannot be less than proportional to so
called finite-order excess entropy of the random text. It is some
mathematical result based on a~line of theorems and one unproved
conjecture.  On the other hand, if Hilberg's hypothesis is true then
the finite-order excess entropy of the text is roughly proportional to
the square root of the text length. The close empirical correspondence
between the nonterminals and the orthographic words allows us to claim
that Hilberg's law implies some lower bound for the vocabulary growth,
i.e.\ some form of Guiraud's law.

The rest of this article fills in the details of the deductions and
empirical observations mentioned in the previous paragraphs:
\begin{itemize}
\item In section \ref{SecHilberg}, we introduce the definitions of
  stationary distribution, block entropy, excess entropy, and infinitary
  distributions. We sketch the history of Hilberg's law
  and the general research of block entropy for natural language
  production.
\item In section \ref{SecGrammars}, we introduce the concepts of
  admissible and irreducible grammars. We also discuss some empirical
  evidence that the shortest admissible grammar matches largely the
  linguistic tokenization for the human text.
\item In section \ref{SecCodes}, we relate block entropy to the
  expected length of irreducible grammar-based codes. Assuming some
  mathematical conjecture, we show that the expected total length of the
  non-initial productions of the shortest grammar cannot be less than
  finite-order excess entropy.
\item In section \ref{SecGuiraud}, we discuss Guiraud's law in detail
  and we argue that Hilberg's law explains it better than the
  assumption of ``intermittent silence''. Some arguments for Hilberg's
  law explanation are: (i) non-randomness of texts, (ii) empirical
  detectability of word boundaries and internal structures, (iii)
  possibility of explaining Menzerath's law, and (iv) significant
  variation of word frequencies across different texts.
\end{itemize}

\section{Excess entropy and Hilberg's law}
\label{SecHilberg}

Let us imagine some infinite sequence of characters, e.g.\
\begin{align}
\label{SomeRoses}
  {\tt the\_rose\_is
    \_a\_hose\_is\_a\_rose\_is\_a\_hose\_is\_a\_rose\_is\_a\_hose...}\,\, ,
\end{align}
where subsequence ${\tt \_a\_rose\_is\_a\_hose\_is}$ is repeated
infinitely to fix our imagination. For such an (infinite) sequence we
can compute the relative frequency of any (finite) string which
appears in that sequence.  

For example, let us define probability $\mathbf{P}({\tt rose})$ as the
relative frequency of string ${\tt rose}$ in the infinite sequence
(\ref{SomeRoses}).  We shall do it in two steps.  Let $a_i$ stand for
the $i$th character of (\ref{SomeRoses}), i.e. $a_{1}={\tt t}$,
$a_{2}={\tt h}$, $a_{3}={\tt e}$, $a_{4}={\tt \_}$, $a_{5}={\tt r}$
etc.  We will write the finite substrings as
$a_{m:n}:=(a_{m},a_{m+1},...,a_{n})$.  The relative frequency
$\mathbf{P}({\tt rose};n)$ of string ${\tt rose}$ in string $a_{1:n}$
is the number of all positions $a_i$, $1\le i\le n$, where string ${\tt
  rose}$ starts divided by $n$.  For any equality relation $\phi$ let
us define $\boole{\phi}=1$ if $\phi$ is true and $\boole{\phi}=0$ if
$\phi$ is false. Thus, $\mathbf{P}({\tt rose};k)$ can be expressed as
\begin{align}
  \mathbf{P}({\tt rose};k):=\frac{1}{k} \sum_{i=1}^k
  \boole{a_{i:i+3}={\tt rose}},
\end{align}
where $:=$ means definition. We have $\mathbf{P}({\tt
  rose};1)=0$, $\mathbf{P}({\tt rose};5)=1/5$, $\mathbf{P}({\tt
  rose};10)=1/10$, $\mathbf{P}({\tt rose};30)=2/30$ and so on.

Let us define probability $\mathbf{P}({\tt rose})$  as the limit
of relative frequencies of string ${\tt rose}$ in the initial substrings of
(\ref{SomeRoses}).  So we will write
\begin{align}
\label{PRose}
\mathbf{P}({\tt rose}):=\lim_{n\rightarrow\infty} \mathbf{P}({\tt rose};n). 
\end{align}
Every $20$th character in sequence (\ref{SomeRoses}) is a~position
where string ${\tt rose}$ starts, so $\mathbf{P}({\tt rose})=1/20$.
Analogically, we can define probability $\mathbf{P}(v)$ for any string
$v$,
\begin{align}
\label{PMises}
\mathbf{P}(v):=\lim_{k\rightarrow\infty} \frac{1}{k}
\sum_{i=1}^k \boole{a_{i:i+\length{v}-1}=v},
\end{align}
where $\length{v}$ is the number of characters in $v$.
Hence, for (\ref{SomeRoses}) we obtain not only $\mathbf{P}({\tt
  t})=0$ (there are no ${\tt t}$'s), $\mathbf{P}({\tt s})=1/5$ (two in
ten characters are ${\tt s}$), and $\mathbf{P}({\tt e})=1/10$ but also
$\mathbf{P}({\tt e\_is\_a})=1/10$, $\mathbf{P}({\tt
  a\_rose\_is\_a\_hose})=1/20$, and $\mathbf{P}({\tt
  a\_rose\_is\_a\_rose})=0$.

Now let us take some general sequence $(a_{1},a_{2},a_{3},...)$.  Let
$\mathbb{V}$ be the finite set of characters that appear in that sequence.
Let $\mathbb{V}^+$ be the set of all finite strings formed by
concatenating the characters in $\mathbb{V}$. For any sequence
$(a_{1},a_{2},a_{3},...)$ such that limit (\ref{PMises}) exists for
each string $v\in\mathbb{V}^+$, probability function $\mathbf{P}$
satisfies relations
\begin{align}
  \label{PConstistency}
  0\le \mathbf{P}(v)\le 1,\quad
  \sum_{a\in\mathbb{V}} \mathbf{P}(a)=1,\quad 
  \sum_{a\in\mathbb{V}} \mathbf{P}(av)= \mathbf{P}(v)=
  \sum_{a\in\mathbb{V}} \mathbf{P}(va).
\end{align}
We will call any function $\mathbf{P}$ satisfying conditions
(\ref{PConstistency}) for all $v\in\mathbb{V}^+$ a~stationary
distribution.\footnote{ Stationary distributions are the distributions
  of stationary stochastic processes \citep{Upper97}. For simplicity,
  we avoid the mathematical terms of stochastic processes, random
  variables and probabilistic spaces
  \citep{Billingsley79,Kallenberg97}.  Since we do not need these
  notions to present the core reasonings, we ignore them to make the
  article as elementary as possible.} It is an open question whether
for any stationary distribution $\mathbf{P}$ exists such
$(a_{1},a_{2},a_{3},...)$ that we have (\ref{PMises}) for all
$v\in\mathbb{V}^+$.

%% Let $f$ be some function which assigns numerical values to strings
%% $v$. We define the expected value of $f$ for strings of length
%% $n$ as
%% \begin{align*}
%%  \sred{f[n]}:=\sum_{v\in\mathbb{V}^n} \mathbf{P}(v)\cdot f(v). 
%% \end{align*}
Let $\mathbb{V}^n$ be the set of all $n$-character long strings.  We
define block entropy $H(n)$ of any stationary distribution
$\mathbf{P}$ as the entropy of strings of length $n$,
\begin{align}
\label{DefEntropy}
  H(n):= 
  -\sum_{v\in\mathbb{V}^n} \mathbf{P}(v) \log_2 \mathbf{P}(v)
  .
\end{align}
We also put $H(0):=0$ for algebraic convenience.

For any stationary distribution $\mathbf{P}$ block entropy $H(n)$ is
a~nonnegative, growing, and concave function of $n$  
\citep{CrutchfieldFeldman01}, i.e., 
\begin{align}
\label{Hproperties}
H(n)\ge 0,\quad 
H'(n)\ge 0,\quad 
H''(n)\le 0,
\end{align}
where 
\begin{align}
  H'(n):=H(n)-H(n-1),\quad 
  H''(n):=H(n)-2H(n-1)+H(n-2).
\end{align}
Because of inequalities (\ref{Hproperties}), we can define entropy
rate as
\begin{align}
\label{DefRate}
h:=\lim_{n\rightarrow\infty}H(n)/n=\lim_{n\rightarrow\infty}H'(n)\ge 0.  
\end{align}

If entropy rate satisfies $h>0$ then $H(n)$ grows almost linearly
against the string length $n$ for very long strings, $H(n)\approx h
n$.  We can ask how fast $H(n)$ approaches $hn$.  The departure of $H(n)$
from the linear growth is known as excess entropy.  

Finite-order excess entropies $E(n)$ are some functions of $H(n)$
and $H(2n)$,
\begin{align}
\label{DefEn}
E(n)&:=2H(n)-H(2n)=-\sum_{k=2}^n (k-1)H''(k)-
\sum_{k=n+1}^{2n} (2n-k+1)H''(k).     
\end{align}
So defined functions are nonnegative and growing, i.e., $E(n)\ge
E(n-1)\ge 0$.
%% Finite-order excess
%% entropies $\bar E(n)$ and $E(n)$ are simple functions of $H(n)$,
%% $H'(n)$ and $H(2n)$,
%% \begin{align}
%% \label{DefbarEn}
%% \bar E(n)&:=H(n)-nH'(n)=-\sum_{k=2}^n (k-1)H''(k), 
%% \\     
%% \label{DefEn}
%% E(n)&:=2H(n)-H(2n)=-\sum_{k=2}^n (k-1)H''(k)-
%% \sum_{k=n+1}^{2n} (2n-k+1)H''(k).     
%% \end{align}
%% Both kinds of finite-order excess entropies are nonnegative and
%% growing, i.e., $\bar E(n)\ge \bar E(n-1)\ge 0$, $E(n)\ge E(n-1)\ge 0$.
%% We also have $\bar E(n)\le E(n)\le \bar E(2n)$. For more discussion on
%% $\bar E(n)$ and $E(n)$, see \citet{CrutchfieldFeldman01}.
\citet{CrutchfieldFeldman01} proved that (total) excess entropy $E$
can be defined equivalently as
\begin{align}
E:=%\lim_{n\rightarrow\infty}\bar E(n)=
\lim_{n\rightarrow\infty}E(n)=
\lim_{n\rightarrow\infty}[H(n)-hn].
\end{align}
We also have inequality
\begin{align}
\label{Control}
  E=-\sum_{k=2}^\infty (k-1)H''(k)\ge -\sum_{k=2}^\infty H''(k)=H(1)-h.
\end{align}

Let $vu$ be the concatenation of strings $v$ and $u$.  We will say
that stationary distribution $\mathbf{P}$ is an IID distribution if
\begin{align}
\label{PRandom}
  \mathbf{P}(vu)=\mathbf{P}(v)\mathbf{P}(u)
\end{align}
for all strings $v,u\in\mathbb{V}^+$.  (IID stands for independent
identically distributed random variables.)  Distributions $\mathbf{P}$
can be IID even for some quite ordered underlying sequences
$(a_{1},a_{2},a_{3},...)$.  For instance, $\mathbf{P}$ given through
(\ref{PMises}) is IID for the sequence of digits of consecutive
natural numbers $(a_{1},a_{2},a_{3},...)=
(1,2,3,4,5,6,7,8,9,1,0,1,1,...)$, which is called Champernowne
sequence \citep{LiVitanyi93}. Anyway, we do not expect that we could
obtain IID distribution $\mathbf{P}$ if we substituted some collection
of human texts for sequence $(a_{1},a_{2},a_{3},...)$.

For any IID distribution $\mathbf{P}$ we have $H(n)=nH(1)$ so $h=H(1)$
and $E=0$. Conversely, if $H(1)-h>0$ or $E>0$, then distribution
$\mathbf{P}$ cannot be IID. For the extreme departures from the IID
case, we have $h=0$ or $E=\infty$. Stationary distributions
exhibiting $h=0$ are called deterministic while the distributions
obeying $E=\infty$ are called infinitary \citep{CrutchfieldFeldman01}.
In appendix \ref{SecMath}, we present some properties of infinitary
distributions which could be important for their possible applications
in quantitative and computational linguistics but which are not so
relevant for the main reasoning of this article.

Let us assume that we could obtain some definite stationary
distribution $\mathbf{P}$ through formula (\ref{PMises}) if we
substituted the infinite concatenation of some human texts for
$(a_{1},a_{2},a_{3},...)$.  We will call such an infinite sequence
$(a_{1},a_{2},a_{3},...)$ natural language production.  Research in
the hypothetical stationary distribution of natural language
production has attracted many scientists.  The first one to work in
this area was \citet{Shannon50}. He tried to estimate block entropy
using the guessing method and assuming some correspondence between
particular instances of English texts and the hypothetical random
English language production. Shannon published some estimates of
$H(n)$ for strings of $n$ consecutive letters, where $n\le 100$.

Shannon was not convinced of any particular asymptotics of block
entropy $H(n)$ for the natural language production
\citep{Hilberg90} but the later researchers in quantitative
linguistics tried to model $H(n)$ by some simple formulae.  For
example, \citet{HoffmanPiotrovskij79} proposed a~model of exponential
convergence,
\begin{align}
\label{HMMH}
H(n)/n=(h_0-h)\exp\kwad{-n/n_0}+h.
\end{align}
\citet{Petrova73} fitted model (\ref{HMMH}) to French language data
and obtained $1/n_0\in (0.24, 0.33)$.  

On the other hand, \citet{Hilberg90} replotted the original plot
of $H(n)$ vs.\ $n$ by \citet{Shannon50} into a~log-log scale and
observed that a simple square-root dependence fits all the data points, 
\begin{align}
\label{HilbergHori}
H(n)\propto n^{\mu},
\quad 
\mu\approx 1/2,
\quad
n\le 100.
\end{align}
For our convenience, we will call Hilberg's law an
algebraic relation which is slightly more general than Hilberg's
original hypothesis (\ref{HilbergHori}).  We will say that Hilberg's
law holds for any stationary distribution $\mathbf{P}$ if only relation
(\ref{HilbergH}) holds with $\mu\approx 1/2$ and $h_\mu>0$ for any $n$.
For such definition, Hilberg's law is independent of any hypothesis on
the particular value of entropy rate $h$ and the constant term $h_0$.

While Shannon estimated block entropy using the guessing method,
Ebeling and his collaborators tried to estimate the asymptotics of
$H(n)$ by counting $n$-tuples in the samples of various symbolic
sequences.  Using improved entropy estimators, the researchers fitted
the general formula (\ref{HilbergH}) with $\mu\approx 1/2$ for natural
language texts and $\mu\approx 1/4$ for classical music transcripts.
For English and German texts $H(n)$ could be safely estimated for
$n\le 30$ characters with $h_0\approx 0$, $h_{\mu}\approx 3.1$ bits
and $h\approx 0.4$ bits \citep{EbelingNicolis92,EbelingPoschel94}.  In
contrast, Shannon's guessing data, reinterpreted by \citet{Hilberg90},
suggest that equation (\ref{HilbergH}) can be extrapolated at least
for $n\le 100$.

It is important to note that the estimation of block entropy $H(n)$
based on the naive estimation of probabilities $\mathbf{P}(v)$ for all
strings $v$ of length $n$ is expensive in the input data.  In order to
estimate the value of $H(n)$, we need a~sample of length about
$2^{H(n)}$ \citep{HerzelSchmittEbeling94}. If we try to make
shortcuts, we assume some particular properties of the unknown
stationary distribution $\mathbf{P}$.  Even \citeauthor{Shannon50}'s
(\citeyear{Shannon50}) guessing method need not give the reliable
estimates of $H(n)$ for the language production if the probabilistic
language model internalized by the experimental subjects differs from
the model estimated from the corpus \citep{BodHayJannedy03,Hug97}.

Let us note that for the block entropy of formula (\ref{HilbergH}),
finite-order excess entropies are
%\bar E(n)\approx h_0 + (1-\mu)h_{\mu}n^{\mu},\quad
\begin{align}
\label{HilbergE}
  E(n)\approx h_0 + (2-2^\mu)h_{\mu}n^{\mu}.
\end{align}
If relations (\ref{HilbergE}) hold with $0<\mu<1$ for any $n$ then the
total excess entropy is $E=\infty$.  Hence, every stationary
distribution exhibiting Hilberg's law is infinitary.  

At the moment, we have no clear idea how one could verify if Hilberg's
law holds for the hypothetical stationary distribution of the language
production. Nevertheless, we can provide some mixed inductive and
deductive arguments that Hilberg's law implies some phenomena that can
be observed in human language.

\section{Words and the shortest grammars}
\label{SecGrammars}

In the following sections, we shall argue that Hilberg's law can
explain some quantitative laws concerning the distribution of word
types in the language production.  Nevertheless, before we can
speak of any distribution of words in a~finite string of phonemes or
letters, we need to delimit the word tokens themselves. If the words
are some objective entities of the language, there should be some
method for identifying the boundaries between the words in
a~sufficiently long string of phoneme or letter tokens even if we
delete the spaces between words and ignore the lexicon.

Let us take some text deprived of spaces, e.g.
\begin{align}
\label{Ex}
 v={\tt shouldawoodchuckchuckifawoodchuckcouldchuckwood}. 
\end{align}
We can express our knowledge of word tokens describing string $v$
by means of a~two-level context-free grammar
\begin{align}
\label{IdealEx}
\begin{array}{rl}
  G&=\klam{
    \begin{array}{ccc}
    \multicolumn{2}{c}{
      b_0\mapsto b_5\,b_1\,b_7\, b_6\, b_2\, b_1\,b_7\, b_4\, b_6\, b_3,} &  
    b_1\mapsto{\tt a}, \\
    b_2\mapsto{\tt if}, & 
    b_3\mapsto {\tt wood}, & 
    b_4\mapsto {\tt could}, \\ 
    b_5\mapsto {\tt should}, & 
    b_6\mapsto {\tt chuck}, & 
    b_7\mapsto {\tt woodchuck} 
    \end{array}
  }
  .
\end{array}
\end{align}
Symbols $b_i$ are called nonterminals. For each $b_i$ there is some
production rule $(b_i\mapsto g_i)\in G$.  On the other hand, the
typewriter-typed symbols, which have no productions rules in the
grammar, will be called terminals. Nonterminal $b_0$ is called the
initial symbol of the grammar. If we recursively substitute
productions $g_i$ for all nonterminals $b_i$ where $(b_i\mapsto
g_i)\in G$, then $b_0$ expands into string $v$ with the requested
tokenization into the words. Namely,
\begin{align*}
 v={\tt \oh{should}\oh{a}\oh{woodchuck}\oh{chuck}
   \oh{if}\oh{a}\oh{woodchuck}\oh{could}\oh{chuck}\oh{wood} 
   },
\end{align*}
where notation $\oh{g}$ means that $G$ contains rule $b_i\mapsto g$
for some $i\neq 0$ \citep{DeMarcken96}. 

Of course, if we were not given any previous knowledge of English
lexicon, we could propose other tokenizations for text (\ref{Ex}).  For
instance, 
\begin{align}
\label{RecCodeEx}
\begin{array}{rl}
  G&=\klam{
    \begin{array}{ccc}
    \multicolumn{2}{c}{b_0\mapsto {\tt sh}  b_1  b_4  b_2
    {\tt if}   b_4  {\tt c}  b_1  b_2  b_3,} &
    b_1\mapsto {\tt ould}, \\ 
    b_2\mapsto {\tt chuck}, & 
    b_3\mapsto {\tt wood}, & 
    b_4\mapsto {\tt a} b_3 b_2
    \end{array}
  }
\end{array}
\end{align}
yields 
\begin{align*}
 v={\tt 
   sh\oh{ould}\oh{a\oh{wood}\oh{chuck}}\oh{chuck}
   if\oh{a\oh{wood}\oh{chuck}}c\oh{ould}\oh{chuck}\oh{wood} 
 }.
\end{align*}
In the extreme, we could define $b_0$ as the entire string $v$ or each
$b_i$, $i\neq 0$, as a~single letter. Since we ignore English lexicon,
we need some purely formal criterion for deciding what grammars $G$
are good for arbitrary strings $v$ and what are not. 

Let us state some formal definitions. Context-free grammar $G$ will
be called a grammar (more precisely, admissible grammar) for string
$v$ \citep[cf.][]{KiefferYang00} if:
\begin{enumerate}
\item For each nonterminal $b_i$ there is exactly one production $g_i$
  such that $(b_i\mapsto g_i)\in G$. 
\item Nonterminal $b_0$ expands into $v$ if we recursively substitute
  productions $g_i$ for all 
  $b_i$.
\end{enumerate}
The set of all admissible grammars for $v$ will be denoted by
$F(v)$. Each grammar $G\in F(v)$ is allowed to produce only one
derivation, which is the finite text $v$ itself. In contrast,
context-free grammars producing a~single infinite derivation are known
as L-systems. 

Some a~priori criterion for deciding which admissible grammars
approximate the correct tokenizations of texts makes use of the
principle of minimum description length
\citep{Rissanen78,LehmanShelat02}.  Define the length $\length{g_i}$
of production $g_i$ as the total number of its terminal and
nonterminal symbols, e.g.\ $\length{{\tt sh} b_1 b_4 b_2 {\tt if} b_4
  {\tt c} b_1 b_2 b_3}=12$ and $\length{{\tt a} b_3 b_2}=3$.
According to the principle of minimum description length, the best
grammar for string $v$ is grammar $G^{\MDL}(v)$ having the minimal
length,
\begin{align}
\label{GMDLDefinition}
G^{\MDL}(v)&:=\mathop{\arg \min}_{G\in F(v)} \length{G},
\end{align}
where the length of a~grammar is the total length of all its productions,
\begin{align}
\label{Glength}
  \length{G}:=\sum_{(b_i\mapsto g_i)\in G} \length{g_i}.
\end{align}
Strictly speaking, there can be more than one grammar having the
minimal length, so object $G^{\MDL}(v)$ is slightly indeterminate.

Grammar $G^{\MDL}(v)$ usually cannot be computed in a~reasonable
amount of time but there is a~multitude of heuristic algorithms which
compute grammars whose lengths approximate $\length{G^{\MDL}(v)}$
\citep{Lehman02,LehmanShelat02}. Various algorithms for computing the
approximations of $G^{\MDL}(v)$ usually perform some kind of local
search on set $F(v)$ and output so called irreducible grammars.
Grammar $G$ is called irreducible \citep[section 3.2]{KiefferYang00}
if:
\begin{enumerate}
\item Each nonterminal expands recursively into a different string
  of terminals.
\item Each nonterminal except for $b_0$ appears at least twice in
  productions $g_i$.
\item There is no string $y$ of $\length{y}\ge 2$ which appears 
  more than once in productions $g_i$.
\end{enumerate}
It can be shown that there is an irreducible grammar
for $v$ whose length equals $\min_{G\in F(v)} \length{G}$. Hence,
we can assume that $G^{\MDL}(v)$ is irreducible.

Various algorithms for computing the irreducible approximations of
$G^{\MDL}(v)$ have been tested empirically on natural language data.
\citet{Wolff80}, \citet{NevillManning96}, and \citet{DeMarcken96}
reported that those algorithms return quite sound representations of
English texts.  The nonterminals of some irreducible approximations of
$G^{\MDL}(v)$ can be interpreted as syllables, morphemes, words, and
fixed phrases.  Some of the heuristic algorithms identify the correct
boundaries of about $90\%$ of orthographic words in the Brown corpus,
in a text deprived of spaces, capitalization, and punctuation
\citep{DeMarcken96}. Here is an example of the computed tokenization
given by \citeauthor{DeMarcken96}:
\begin{center}
$
\begin{array}[c]{c}
{\tt 
\oh{ \oh{f\oh{or}} \oh{ \oh{t\oh{he}} \oh{
      \oh{\oh{p\oh{ur}}\oh{\oh{\oh{po}s}e}} \oh{of} }}}
\oh{\oh{\oh{ma\oh{in}}\oh{ta\oh{in}}}\oh{\oh{in}g}}
\oh{\oh{\oh{in}\oh{t\oh{er}}}\oh{\oh{n\oh{a\oh{t\oh{i\oh{on}}}}}\oh{al}}}
}
\\
{\tt 
\oh{\oh{pe}\oh{a\oh{ce}}} \oh{\oh{an}d}
\oh{\oh{\oh{p\oh{ro}}\oh{\oh{mo}t}}\oh{\oh{in}g}}
\oh{t\oh{he}}
\oh{\oh{adv\oh{a\oh{n\oh{ce}}}}\oh{\oh{\oh{me}n}t}} \oh{
  \oh{of} \oh{a\oh{ll}} }
}
\\
{\tt 
 \oh{\oh{pe}\oh{op}\oh{le}} \oh{\oh{ \oh{t\oh{he}} \oh{
      \oh{\oh{un}\oh{it}\oh{ed}} \oh{\oh{\oh{st\oh{at}}e}s} }}
  \oh{ \oh{of} \oh{a\oh{me}\oh{r\oh{ic}}a} }}
\oh{\oh{\oh{jo}\oh{in}}\oh{ed}}
}
\\
{\tt 
\oh{in} \oh{f\oh{o\oh{un}d}} \oh{\oh{in}g} \oh{ \oh{t\oh{he}}
  \oh{ \oh{\oh{un}\oh{it}\oh{ed}}
    \oh{\oh{n\oh{a\oh{t\oh{i\oh{on}}}}}s} }}
}
.  
\end{array}
$
\end{center} 

The results of the automatic tokenization are especially impressive
for strongly isolating languages, such as English and Chinese
\citep{DeMarcken96}. The same algorithms need not be so effective for
highly inflective languages, where numerous orthographic alternations
occur within the morphological stems (e.g.\ for Polish).  The pursuit
for better tokenization algorithms cannot be separated from the quest
for the data compression algorithms which identify the inflectional
paradigms \citep{Goldsmith01} or the abstract phrase syntax structures
\citep{NowakPlotkinJansen00}.

\section{The shortest grammar and excess entropy}
\label{SecCodes}

Let us denote the set of the non-initial rules of grammar $G$ as
$G_0:=G\setminus \klam{b_0\mapsto g_0}$, where $A\setminus B$ is the
difference of sets $A$ and $B$.  We will call $G_0$ the vocabulary of
$G$.  The length of the vocabulary is defined as
\begin{align}
  \length{G_0}:=\sum_{(b_i\mapsto g_i)\in G_0}  
  \length{g_i}=\length{G} -\length{g_0}.
\end{align}
We use notation
$\length{G_0^{\MDL}(v)}:=\length{G^{\MDL}(v)} -\length{g_0^{\MDL}(v)}$
respectively. 

If the average length of the word-like productions $g_i$, $i\neq 0$,
does not depend significantly on the text then we may suppose that
$G_0^{\MDL}(v)$ is proportional to the number of word types in text
$v$.  In fact, we can observe an analog of Guiraud's law
(\ref{Guiraud}). If we look at the data published by \citet[figure
3.12 (b), p.\ 69]{NevillManning96}, we can observe empirical
proportionality
\begin{align}
\label{NevillVocabulary}
 \length{G_0^{\SEQUITUR}(v)}\propto (\length{v})^\alpha, 
\end{align}
where $1/2<\alpha<1$ and $G_0^{\SEQUITUR}(v)$ is some approximation of
$G_0^{\MDL}(v)$ computed by the algorithm called $\SEQUITUR$.
%(a) rules in the grammar (b) symbols in the grammar (c) vocabulary
%size in the input vs.\ input characters

In this section, we would like to present some general theoretical
result.  We shall relate the length of $G_0^{\MDL}(v)$ to the
finite-order excess entropy.  It is well known that there are intimate
relations between block entropy and the expected lengths of some codes
used in data compression.  In particular, \citet{KiefferYang00}
discuss the concept of grammar-based codes, which represent strings
$v\in\mathbb{V}^+$ as uniquely decodable binary strings
$C(v)\in\klam{0,1}^+$ by the mediation of the admissible grammars.

Let $F=\bigcup_{v\in\mathbb{V}^+} F(v)$ be the set of admissible
grammars for all strings.  Function
$C:\mathbb{V}^+\rightarrow\klam{0,1}^+$ is called a~grammar-based code
if
\begin{align}
\label{GrammarBased}
C(v)=\mathbf{B}(G^C(v)),  
\end{align}
where grammar transform $G^C$ computes grammar $G^C(v)\in F(v)$ and
grammar encoder $\mathbf{B}$ represents any grammar $G\in F$ as
a~unique binary string $\mathbf{B}(G)\in\klam{0,1}^+$.
% It is also assumed that the subset of binary strings
% $\mathbf{B}(F):=\klam{\mathbf{B}(G): G\in F}$ is prefix-free, i.e.\
% no string $w\in \mathbf{B}(F)$ is an initial substring of another
% string $w'\in \mathbf{B}(F)$.

Let us introduce the expected length of code $C$ for the strings
of length $n$ drawn from stationary distribution $\mathbf{P}$,
\begin{align}
\label{ExpectedMDL} 
 H^C(n)&:=
 \sum_{v\in\mathbb{V}^n} \mathbf{P}(v)\cdot \length{C(v)}.
\end{align}
% By the assumption that set $\mathbf{B}(F)$ is prefix-free we have
% \begin{align}
% \label{ChannelInequality}
%   H^C(n) &\ge H(n),
% \end{align}
% where $H(n)$ is the block entropy of $\mathbf{P}$ \citep[sections 5.2,
% 5.3]{CoverThomas91}. 
Code $C$ is called universal (more precisely, weakly minimax
universal) if
\begin{align}
\label{ChannelInequality}
  H^C(n) &\ge H(n),
\\
\label{RateEquality}
\lim_{n\rightarrow\infty}H^C(n)/n &=\lim_{n\rightarrow\infty}H(n)/n
\end{align}
for any stationary distribution $\mathbf{P}$.  See \citet[sections
5.1--6 and 12.10]{CoverThomas91} for a~general background in
information and coding theory.

Additionally, let us call $C$ an irreducible code if for each input
string $v\in\mathbb{V}^+$, grammar $G^C(v)$ is irreducible.
\citet[theorem 8]{KiefferYang00} prove the following result:
\begin{theorem}
\label{theoEncoder}
There exists such grammar encoder $\mathbf{B}$ that any irreducible
code of form (\ref{GrammarBased}) is weakly minimax universal.
\end{theorem}
\noindent
It is a~very strong and profound theorem. In particular, code
$\MDL(v):=\mathbf{B}(G^{\MDL}(v))$ is universal since the shortest grammar
$G^{\MDL}(v)$ is irreducible. Theorem \ref{theoEncoder} can be used to
prove universality of the modified $\SEQUITUR$ code by
\citeauthor{NevillManning96} \citep[section 6.2]{KiefferYang00}.
Universality of the famous Lempel-Ziv code, however, is proved
differently since it is not an irreducible code and it uses
a~different grammar encoder \citep[section 12.10]{CoverThomas91}.

It has been checked empirically that codes whose grammars are shorter
usually enjoy shorter lengths.  For instance, \citet{Grassberger02}
compressed 135 GB of English text and obtained compression rates (in
bits per character) $\length{\LZ(v)}/\length{v} \approx 2.6$ for
Lempel-Ziv code $\LZ$ and $\length{\NSRPS(v)}/\length{v} \approx 1.8$
for some heuristic irreducible code $\NSRPS$.  Other researchers
reported comparable results \citep{DeMarcken96}.

By analogy to definition (\ref{DefEn}) of finite-order excess entropy
$E(n)$, let us introduce the expected excess code length
\begin{align}
  \label{DefThn}
  E^C(n)&:= 2H^C(n)-H^C(2n)
  \nonumber
  \\
  &\,\,= \sum_{v,u\in\mathbb{V}^n} \mathbf{P}(vu) 
  \kwad{\length{C(v)} +  \length{C(u)} - \length{C(vu)}}
  .
\end{align}
\begin{theorem}
  \label{theoDiffThEbar}
  For any weakly minimax universal code $C$ inequality
\begin{align}
\label{DiffThEbar}
  E^C(n)\ge E(n)
\end{align}
is true for infinitely many $n$. (See appendix \ref{SecProof} for the proof.)
\end{theorem}
\noindent
Inequality (\ref{DiffThEbar}) is valid in particular for $C=\MDL$ or
for any irreducible code.

Now, we shall link the expected excess code length $E^{\MDL}(n)$ with
the length of $\MDL$ vocabulary. Let $L^m(v):=\length{G^{\MDL}(v)}$ be
the length of the shortest grammar and
$L_0^m(v):=\length{G_0^{\MDL}(v)}$ be the length of its vocabulary.
Define $L^{>1}(v)$ as the maximal length of a~string which appears in
string $v$ at least twice.
\begin{theorem}
  \label{theoVocabulary} 
  We have inequalities 
  \begin{align}
    \label{ReZero}
    L^m(v)&\le \length{v},
    \\
    \label{ReOne}
    L^m(v),L^m(u)&\le L^m(vu) + L^{>1}(vu),
    \\
    \label{ReBoth}
    0 \le L^m(v)+L^m(u)-L^m(vu)&\le L_0^m(vu) +
    L^{>1}(vu).
  \end{align}
  (See appendix \ref{SecProof} for the proof.)
\end{theorem}
\noindent

Inequality (\ref{ReBoth}) states that the vocabulary length for the
shortest grammar cannot be roughly less than the excess length of the
shortest grammar.  In a~slightly heuristic reasoning, we shall argue
that the excess length of the shortest grammar multiplied by a~slowly
growing function cannot be less than the excess length of code $\MDL$.
In order to do it we need some pretty strong symmetrical bound for the
length of code $\MDL$ in terms of the length of the shortest grammar.

It is known that function $\mathbf{B}$ of
Theorem \ref{theoEncoder} satisfies $\len \mathbf{B}(G)\le
\gamma(\length{G})$, where $\gamma(n):= n\cdot (c +\log n)$ for some
constant $c$ \citep[section 4]{KiefferYang00}.  The following
symmetrical bound for code $\MDL$ seems probable:
\begin{conjecture}
  \label{conjBound} 
  There is inequality
  \begin{align}
    \abs{\length{\MDL(v)}-\gamma(L^m(v))}\le f_2(L^m(v)),
  \end{align}
  where $\gamma(n):= n\cdot f_1(n)$ and functions $f_i\ge 0$ satisfy
  $0\le f_i(n+1)-f_i(n)\le c_i/n$ for some constants $c_i$.
\end{conjecture}

Now we can give a~bound for the excess length of code $\MDL$ in terms of
the excess length of the shortest grammar.
\begin{theorem}
  \label{theoIfBound}
  If Conjecture \ref{conjBound} is true then
  \begin{align}
  &\length{\MDL(v)} 
  + \length{\MDL(u)} 
  - \length{\MDL(vu)}
  \nonumber
  \\
    \label{IfBound}
  &\quad\le 
  \kwad{L^m(v)+L^m(u)-L^m(vu) + d_1}\kwad{d_2 + c_1  \log \length{vu}
    +c_1 \frac{L^{>1}(vu)}{L^m(vu)}}
  + c_1 L^{>1}(vu),
\end{align}
where $d_1=3c_2/c_1$ and $d_2=\max (f_1(1),f_2(1)c_1/c_2)$. (See
appendix \ref{SecProof} for the proof.)
\end{theorem}

Recall that $H^{\MDL}(n)/n=\sum_{v\in\mathbb{V}^n} \mathbf{P}(v)\cdot
L^m(v)/\length{v}$ approaches entropy rate $h$ for
$n\rightarrow\infty$ by Theorem \ref{theoEncoder}.  We may speculate
that $h>0$ for the language production. Let us assume a~stronger
statement, namely, that
\begin{align}
 \length{v} \le d_3 L^m(v)
\end{align}
for some constant $d_3$ and (almost) every human text $v$.  On the other hand,
notice that $L^{>1}(v)\le \length{v}$ follows by definition of
$L^{>1}(v)$. By these two inequalities, we have $L^{>1}(vu)/L^m(vu)\le
d_3$.  Combining the latter with (\ref{IfBound}) and (\ref{ReBoth})
gives
\begin{align}
   \length{\MDL(v)} 
  &+ \length{\MDL(u)} 
  - \length{\MDL(vu)}
  \nonumber
  \\
    \label{ThenBound}
  &\le 
  \kwad{L_0^m(vu) +L^{>1}(vu) +d_1}\kwad{d_4 + c_1 \log
    \length{vu}}, 
\end{align}
where $d_4:=d_2+ c_1(d_3+1)$. Averaging (\ref{ThenBound}) 
 with $\mathbf{P}(vu)$ for $v,u\in\mathbb{V}^n$, we obtain
 \begin{align}
  \label{SolMDLbar}
   \kwad{ L_0^m[2n] + L^{>1}[2n]+ d_1}\kwad{d_4 + c_1
     \log (2n)} \ge E^{\MDL}(n),
 \end{align}
where
\begin{align}
\label{ExpectedGMDL} 
 L_0^m[n]:=
 \sum_{v\in\mathbb{V}^n} \mathbf{P}(v)\cdot \length{L_0^m(v)},
 \quad
 L^{>1}[n]:=
 \sum_{v\in\mathbb{V}^n} \mathbf{P}(v)\cdot \length{L^{>1}(v)}.
\end{align}

By inequality (\ref{SolMDLbar}) and Theorem \ref{theoDiffThEbar}, 
we also have
\begin{align}
\label{DiffGMDLEbar}
\kwad{ L_0^m[2n] + L^{>1}[2n]+ d_1}
\kwad{d_4 + c_1\log (2n)}\ge E(n)
\end{align}
for infinitely many $n$. In particular, if stationary distribution
$\mathbf{P}$ obeys Hilberg's law (\ref{HilbergH}) then inequality
\begin{align}
 \label{GMDLHilberg}
 L_0^m[n] + L^{>1}[n] \ge \const\cdot n^{\mu}/\log n
\end{align}
holds for infinitely many $n$ by equation (\ref{HilbergE}).

\section{Hilberg's law and Guiraud's law}
\label{SecGuiraud}

In this section, we would like to make the final step in deriving
Guiraud's law from relation (\ref{GMDLHilberg}). First, let us have
a~closer look at Guiraud's and Zipf's laws.  It is widely-known that
if Zipf's law (\ref{Zipf}) holds with the same $B$ for all $N$
then Guiraud's law (\ref{Guiraud}) is satisfied with $\rho=1/B$ for
large $N$, cf.\ \citet[section 3.2]{Kornai02} or
\citet{FerrerSole01b}. 

In fact, the number of word types $V$ and the number of word
tokens $N$ can be computed given the word frequencies,
\begin{align}
V&=\sum_{w:\,c(w)>0} 1, \quad
N=\sum_{w:\,c(w)>0} c(w),
\end{align}
so any relation between $V$ and $N$ is a~function of the exact
distribution of frequencies $c(w)$. The converse is not true.  In
general, frequency $c(w)$ cannot be computed given only $w$, $V$, and
$N$ since different texts usually have  different keywords.
Still, we may seek for hypothetical derivations of formula
(\ref{Zipf}) given formula (\ref{Guiraud}) and some additional
assumptions.

One could ask if Guiraud's law or Zipf's law do hold with the same
$\rho$ or $B$ for texts of various size and origin. The answer is
complex. For instance, \citet[section 2.5]{Kornai02} discusses
Guiraud's law extensively and according to the plot in his article
value $\rho\approx 0.75$ holds perfectly for samples of sizes
$N\in\kwad{1.4\cdot 10^5,1.8\cdot 10^7}$ drawn from San Jose Mercury
News corpus.  Such value of $\rho$ would correspond to $B\approx 1.33$
if formula (\ref{Zipf}) with constant $B$ held for all word ranks.
Nevertheless, if we investigate the rank-frequency plot for so large
collections of texts, we encounter a~different regularity.

\citet{FerrerSole01b} discovered that parameter $B$ in formula
(\ref{Zipf}) depends on word rank $r(w)$. For multi-author corpora 
there are two regimes where $B$ is almost constant.  Namely, we have
\begin{align}
\label{TwoRegimes}
  B=
  \begin{cases}
    B_1, & 0\le r(w)\le R_1,
    \\
    B_2, & R_1\le r(w),
  \end{cases}
\end{align}
where $B_2<B_1\approx 1$.  Let us note that for sufficiently short
text collections (those with $V < R_1$) only one of two regimes can be
observed. For single-author corpora and $r(w)\ge R_1$, we have an
exponential decay of $c(w)$ rather than a~power-law.

In another case of some multi-author collection of English texts
counting $1.8\cdot 10^8$ word tokens, \citet{MontemurroZanette02}
reported $B_1\approx 1$, $B_2\approx 2.3$ and $R_1\approx 6000$. The
investigated collection is only 10 times larger than SJMN corpus
surveyed by Kornai.  If formula (\ref{Zipf}) with constant $B\approx
2.3$ held for all word ranks then we would have Guiraud's law
(\ref{Guiraud}) with $\rho\approx 0.43$.  Anyway, if there are two
regimes of $B$, like in (\ref{TwoRegimes}), then we could obtain
Guiraud's law (\ref{Guiraud}) with $\rho\approx 0.75$ for all $N$ if
also parameter $R_1$ depends on the text length $N$.
Until we have more experimental data on the dependence between $N$ and
$R_1$, we can be only sure that there is inequality
\begin{align}
\label{GuiraudH}
  V\ge \const\cdot N^{0.43}.
\end{align}

Let $V(v)$ be the number of orthographic word types in text $v$
and $N(v)$---the number of orthographic word tokens therein.  If
we assume that the mean length of the word tokens in text $v$ does not
change substantially with $v$ then text length $N(v)$ measured in
orthographic words is proportional to text length $\length{v}$
measured in phonemes or letters,
\begin{align}
  \label{Npropto}
  N(v)\propto \length{v}.
\end{align}

In view of section \ref{SecGrammars}, we may suppose that the number
of orthographic word types $V(v)$ is proportional to the number
of the production rules in the shortest grammar $G^{\MDL}(v)$, cf.\
\citet[figure 3.12 (c) vs.\ (a), p.\ 69]{NevillManning96}.  If the
mean length of the non-initial productions does not change
substantially against $v$ then the number of the rules is proportional
to length $L_0^{m}(v)$ of the vocabulary of the shortest grammar
$G^{\MDL}(v)$, cf.\ \citet[figure 3.12 (a) vs.\ (b), p.\
69]{NevillManning96}.  Resuming, we would have proportionality
\begin{align}
  \label{Vpropto}
  V(v)\propto L_0^{m}(v).
\end{align}
%\citet[figure 3.12 (b), p.\ 69, unfortunately it is not a~log-log
% plot]{NevillManning96}.
%(a) rules in the grammar (b) symbols in the grammar (c) vocabulary
%size in the input vs.\ input characters

Assuming relations (\ref{Npropto}) and (\ref{Vpropto}),
we can restate Guiraud's law (\ref{GuiraudH}) as
\begin{align}
 \label{GMDLGuiraudH}
  L_0^{m}(v)\ge \const\cdot (\length{v})^{0.43},
\end{align}
which resembles relation (\ref{NevillVocabulary}) reported by
\citeauthor{NevillManning96}. Except for the effects of averaging and
the negligible length $L^{>1}(v)$ of the longest substring appearing
more than once, inequality (\ref{GMDLGuiraudH}) is implied by
inequality (\ref{GMDLHilberg}) with the very rough estimate
$\mu\approx 1/2$ done by Hilberg. We could say that Hilberg's law can
be some explanation of Guiraud's law. Let us discuss the plausibility
of such explanation.

Zipf's law is often understood as a~specific algebraic relationship
between the counts and ranks of various objects---not necessarily words.
In such generalization, Zipf's law is observed also out of the
linguistic domain, e.g.\ in income distribution \citep{Pareto97}.
We do not know if one can find a~general explanation of Zipf's law
both in linguistic and non-linguistic contexts.  Explaining Zipf's
law in the purely linguistic context seems somehow easier. One needs ``only''
to assign some reasonable relative frequency $\mathbf{P}(v)$ to
every string $v$ of phonemes and then to
define how any finite string $v$ should be cut into words.  The
existence or nonexistence of relation (\ref{Zipf}) should follow by
pure mathematical deduction from these two assumptions.

That idea inspired \citet{Mandelbrot53} to formulate some classical
explanation of Zipf's law. His assumptions are:
\begin{enumerate}
\item Stationary distribution $\mathbf{P}$ is an IID distribution,
  i.e.\ it satisfies (\ref{PRandom}).
\item Set $\mathbb{V}$ of atomic symbols is the set of 
  phonemes and spaces. The word tokens in any text are defined as the
  space-to-space strings of phonemes.
\end{enumerate}
Given these assumptions Mandelbrot derived Zipf's law for
space-to-space words and hence Guiraud's law can be inferred as well.
In fact, Mandelbrot did not discuss Guiraud's law but, as we have
said, Zipf's law does imply Guiraud's law automatically.  Mandelbrot's
explanation assuming the existence of ``intermittent silences'' was
quoted or rediscovered by many researchers, e.g.\ by
\citet{Belevitch56}, \citet{Miller57}, \citet{BellClearyWitten90} and
\citet{Li92}. There is some historical summary of that literature done
by \citet{Li98}.

Although Mandelbrot's explanation of Zipf's and Guiraud's laws earned
some popularity among natural scientists, we should stress that both
of its assumptions are false with respect to the intended application
to natural language. First, we would object to modeling human language
production by an IID distribution.  Second, Mandelbrot's definition of
word is biased by the spelling conventions of the most popular
alphabetic scripts which use blank spaces to separate words.  No
regular ``intermittent silences'' appear in the spoken versions of the
corresponding ethnic languages \citep{Jelinek97}. That phenomenon is
a~challenge for automatic speech recognition and it motivated some
interest in the shortest admissible grammars as a~means for restoring
the boundaries between the words \citep{DeMarcken96}.

In this article, we present another explanation of Guiraud's law.  Our
assumptions are: 
\begin{enumerate}
\item Stationary distribution $\mathbf{P}$ exhibits Hilberg's law
  (\ref{HilbergH}) for all $n$.
\item We may assume that $\mathbb{V}$ is a~set of 
  phonemes only.  The word tokens in any text are defined as the
  nonterminal tokens of the shortest admissible grammar.
\end{enumerate}
We think that the derivation of Guiraud's law based on Hilberg's law
is better linguistically justified than the classical explanation
by Mandelbrot. There are several reasons for that claim:
\begin{enumerate}
\item The new explanation assumes that human narration
  exhibits strong probabilistic dependence, it is not a~IID distribution.
  In appendix \ref{SecMath}, we recall that no infinitary
  distribution $\mathbf{P}$ can be modeled by a~stationary hidden
  Markov chain with a~finite number of hidden states. This fact can
  have some important implications for computational linguistics
  \citep{Jelinek97}.
\item The new explanation does not assume the pre-existence of spaces
  between the words in the natural language production. Children can
  learn the correct tokenization of speech into the words even if they
  do not know yet what the words are.
\item Space-to-space words for the IID distributions do not have any
  definite internal structure. It is no longer true for the new
  explanation.  The nonterminals of the shortest grammar
  exhibit the internal structure of recursive rule
  productions.  Such nonterminals have well-defined parts.  Without
  any change of the model, we can speak not only of Guiraud's and
  Zipf's laws for the words but we can also discuss laws which relate
  words to their elements. Some example of the latter is Menzerath's
  law, which states that the longer the word is the shorter its
  constituents are \citep{Menzerath28,Altmann80}.  By means of the
  grammar-based codes one can define the structure of word-like
  objects and investigate many quantitative linguistic laws not only
  for the language production but also for any other stationary
  distributions.
\item Stationary distribution is called ergodic (roughly) if the
  relative frequency of any fixed word does not vary significantly
  across different texts. By some theorem, every IID distribution is
  ergodic \citep[chapter 4]{Debowski05}.  Nevertheless, empirical studies do
  not corroborate Mandelbrot's assumption that language production
  $\mathbf{P}$ is ergodic. The mere existence of concept ``the
  keywords of the text'' reflects the fact that different texts use
  different vocabularies systematically.  Words, once they appear in
  some text, tend to reappear. Let us stress that some significant
  variation of the word frequencies \textit{can} be modelled by
  non-ergodic stationary distributions. Many non-ergodic stationary
  distributions are infinitary \citep[chapters 4 and 5]{Debowski05}, see also
  appendix \ref{SecMath}. It is an interesting question whether
  Hilberg's law (\ref{HilbergH}) implies non-ergodicity of stationary
  distribution $\mathbf{P}$. Some further discussion of Hilberg's law
  and non-ergodic distributions could give us insight where to seek
  general quantitative laws in the intertext variability of language.
  Any such laws would be of great importance to computational
  linguistics as well.
\end{enumerate}

\section{Conclusions}
\label{SecConclusions}

In this article, we have discussed some implications of
\citeauthor{Hilberg90}'s (\citeyear{Hilberg90}) hypothesis on the
entropy of natural language production. That hypothesis states that
finite-order excess entropy $E(n)$ of the $n$-letter strings is
proportional to the square root of $n$. So far, the proportionality
has been roughly verified only for $n\le 50$.  On the other hand, we
have argued that Hilberg's hypothesis, when extrapolated to $n$ of the
text length magnitude, provides a~better explanation of Guiraud's law
than the classical explanation based on the existence of
``intermittent silences'' \citep{Mandelbrot53}.

The new explanation is based on two points.  First, we observe that
the tokenization of a~text into orthographic words and their morphemes
matches largely the production rules of the shortest admissible
grammar for the text.  Second, we use some partially heuristic, but
largely deductive, mathematical reasoning to argue that the length of
the non-initial production rules of the shortest grammar cannot be less
than finite-order excess entropy.

In the future research, the rough match of the
linguistically-motivated tokenizations and the tokenizations given by
the shortest grammars should be surveyed as one of the fundamental
problems of quantitative linguistics.  One should survey Zipf's,
Guiraud's, and Menzerath's laws for the nonterminals of the admissible
grammars and the orthographic words simultaneously across a~large
range of text sizes and languages.  Proportionalities (\ref{Npropto})
and (\ref{Vpropto}) should be verified as well.

It seems that the existence of a~rich formal structure in the natural
language production is reflected by its high total excess entropy $E$
rather than by simply positive entropy gain $H(1)-h$.  We think that
the further discussion of Hilberg's hypothesis can improve the quality
of statistical language models both in quantitative and computational
linguistics, see appendix \ref{SecMath} and our doctoral
dissertation \citep{Debowski05}.

Since the shortest admissible grammars reproduce also the internal
structure of words, the behavior of excess entropy might be linked not
only with Guiraud's and Zipf's laws but also with Menzerath's law. The
shortest grammars can be used as the \textit{definition} of words and
their constituents in any symbolic string \citep{NevillManning96}.
Adopting such a~definition, empirical researchers can survey the form of
Guiraud's, Zipf's, and Menzerath's laws also in the non-linguistic
symbolic data (such as DNA).  Last but not least, mathematicians
can prove some rigorous theorems.

%\begin{flushleft}
  \bibliography{books,ql,nlp,ai,neuro,mine,tcs}
%\end{flushleft}

\appendix

\section{Proofs}
\label{SecProof}

\begin{proof*}{\ref{theoDiffThEbar}}
For any function $f$ we have identity
\begin{align}
  \sum_{k=0}^{m-1} 
\kwad{2f(2^k n)-f(2^{k+1} n)}\cdot\frac{1}{2^{k+1}}=
f(n)-\frac{f(2^m n)}{2^m n}\cdot n
\end{align}
for each finite $m$. Hence, if (\ref{RateEquality}) is true then we
obtain
\begin{align}
  H(n)-hn &= \sum_{k=0}^\infty 
\kwad{2H(2^k n)-H(2^{k+1} n)}\cdot\frac{1}{2^{k+1}}=
\sum_{k=0}^\infty \frac{E(2^k n)}{2^{k+1}}
,
\\
  H^C(n)-hn &= \sum_{k=0}^\infty 
\kwad{2H^C(2^k n)-H^C(2^{k+1} n)}\cdot\frac{1}{2^{k+1}}=
\sum_{k=0}^\infty \frac{E^C(2^k n)}{2^{k+1}}
.
\end{align}
Because of inequality (\ref{ChannelInequality}), we have
$H(n)-hn\le H^C(n)-hn$ so 
\begin{align}
\label{ThEbarSeries}
  \sum_{k=0}^\infty \frac{E(2^k n)}{2^{k+1}}\le 
  \sum_{k=0}^\infty \frac{E^C(2^k n)}{2^{k+1}}.
\end{align}
If we put $n=2^p M$ with any $p$ and some fixed $M$ then
(\ref{ThEbarSeries}) yields
\begin{align}
\label{ThEbarSeries2}
\sum_{k=p}^\infty \frac{E^C(2^k M)-E(2^k M)}{2^{k+1}}\ge
0.
\end{align}

Assume that $E^C(2^k M)-E(2^k M)\ge 0$ holds only for finitely many
$k$. Then we would have $E^C(2^k M)-E(2^k M)< 0$ for all $k\ge p$
and some $p$. Hence, we would have
\begin{align}
\label{ThEbarSeries3}
\sum_{k=p}^\infty \frac{E^C(2^k M)-E(2^k M)}{2^{k+1}}<
0.
\end{align}
Since (\ref{ThEbarSeries3}) stays in contradiction with
(\ref{ThEbarSeries2}), our assumption that $E^C(2^k M)-E(2^k M)\ge 0$
only for finitely many $k$ was false. We must have $E^C(2^k
M)-E(2^k M)\ge 0$ for infinitely many $k$, and this is exactly
inequality (\ref{DiffThEbar}) which we were to prove.
\end{proof*}

%% It is worth noting that in order to prove (\ref{DiffThEbar}), we do
%% not need to assume that $E^C(n)\ge 0$ holds for all $n$, i.e., that
%% we have subadditivity (\ref{Subadditivity}) for all $n_1$ and $n_2$.
  
\begin{proof*}{\ref{theoVocabulary}}
In order to prove (\ref{ReZero}), notice that $G=\klam{b_0\mapsto
  v}$ is a~grammar for $v$. Its length satisfies
$\length{v}=\length{G}\le \length{G^{\MDL}(v)}$ by (\ref{Glength})
and (\ref{GMDLDefinition}).

Now, let us prove (\ref{ReOne}) and (\ref{ReBoth}).  Since vocabulary
$G_0^{\MDL}(vu)$ cannot beat vocabularies $G_0^{\MDL}(v)$ and
$G_0^{\MDL}(u)$ in the efficient representation of any strings $v$ and
$u$ respectively, we observe inequalities
\begin{align}
  \label{Left}
  \length{G^{\MDL}(v)} 
  &\le
  \length{g_L} +\length{G_0^{\MDL}(vu)},
  \\
  \label{Right}
  \length{G^{\MDL}(u)} 
  &\le
  \length{g_R} +\length{G_0^{\MDL}(vu)},
\end{align}
where $G_0^{\MDL}(vu)\cup \klam{b_0\mapsto g_L}$ and
$G_0^{\MDL}(vu)\cup \klam{b_0\mapsto g_R}$ are some grammars for $v$
and $u$ respectively. Analogically,
\begin{align}
  \label{Inverse}
  \length{G^{\MDL}(vu)} 
  &\le \length{G^{\MDL}(v)} + \length{G^{\MDL}(u)} 
\end{align}
since $G_0^{\MDL}(v)\cup G_0^{\MDL}(u)\cup \klam{b_0\mapsto
  g_0^{\MDL}(v) g_0^{\MDL}(u)}$ is a~grammar for
$vu$.

Assume that $g_L$ and $g_R$ are obtained by splitting the initial
production $g_0^{\MDL}(vu)$ into two parts and recursively expanding
the nonterminal at the border if necessary.  That is, we have either
$g_Lg_R=g_0^{\MDL}(vu)$ or $g_L=y_Lx_L$, $g_R=x_Ry_R$, and
$g_0^{\MDL}(vu)=y_Lb_iy_R$, where nonterminal $b_i$ expands
recursively into string $x_Lx_R\in\mathbb{V}^+$. Grammar
$G^{\MDL}(vu)$ is irreducible so we must have $\length{x_Lx_R}\le
L^{>1}(vu)$, where $L^{>1}(vu)$ is the maximal length of a~string
which appears in string $vu$ at least twice. Thus,
\begin{align}
\label{Sum}
  \abs{\length{g_L}+\length{g_R}- \length{g_0^{\MDL}(vu)}}\le L^{>1}(vu).
\end{align}
By (\ref{Sum}), adding (\ref{Left}) and (\ref{Right}) yields
\begin{align}
  \length{G^{\MDL}(v)}  + \length{G^{\MDL}(u)}
  &\le \length{g_0^{\MDL}(vu)} + 2\length{G_0^{\MDL}(vu)} + L^{>1}(vu)
  \nonumber
  \\
  \label{Both}
  &= \length{G^{\MDL}(vu)} + \length{G_0^{\MDL}(vu)} + L^{>1}(vu).
\end{align}
In fact, we can rewrite (\ref{Both}) and (\ref{Inverse}) as
(\ref{ReBoth}).  By (\ref{Sum}), we also have
$\length{g_L},\length{g_R}\le \length{g_0^{\MDL}(vu)}+ L^{>1}(vu)$.
Inserting these two inequalities into (\ref{Left}) and (\ref{Right})
respectively yields (\ref{ReOne}).
\end{proof*}

\begin{proof*}{\ref{theoIfBound}}
  According to Conjecture \ref{conjBound}, we have
  \begin{align}
    \label{BoundI}
  &\length{\MDL(v)} 
  + \length{\MDL(u)} 
  - \length{\MDL(vu)}
  \nonumber
  \\
  &\quad\le
  \gamma(L^m(v))+ \gamma(L^m(u))-\gamma(L^m(vu))
  + f_2(L^m(v))+f_2(L^m(u))+f_2(L^m(vu))    
  \end{align}
  By $0\le f_i(n+1)-f_i(n)\le c_i/n$
  and  (\ref{ReOne}), there is
  \begin{align}
    \label{BoundL}
     f_i(n)&\le f_i(1)+\sum_{k=2}^n c_i/k<f_i(1) +c_i\log n,
     \\
     f_i(L^m(v))
     &\le 
     f_i(L^m(vu)) + c_i L^{>1}(vu)/L^m(vu).
    \end{align}
  Hence by (\ref{ReZero}),
  \begin{align}
    \label{BoundG}
    \gamma(L^m(v))&+ \gamma(L^m(u))-\gamma(L^m(vu))
    \nonumber
    \\
    &\le 
    \kwad{L^m(v)+L^m(u)-L^m(vu)} f_1(L^m(vu)) 
    + c_1 \kwad{L^m(v)+L^m(u)}\frac{L^{>1}(vu)}{L^m(vu)}
    \nonumber
    \\
    &=
    \kwad{L^m(v)+L^m(u)-L^m(vu)} 
    \kwad{f_1(L^m(vu)) + c_1\frac{L^{>1}(vu)}{L^m(vu)}}
    + c_1 L^{>1}(vu)
    \nonumber
    \\
    &\le 
    \kwad{L^m(v)+L^m(u)-L^m(vu)} 
    \kwad{f_1(\length{vu}) + c_1 \frac{L^{>1}(vu)}{L^m(vu)}}
    + c_1 L^{>1}(vu)
    .
  \end{align}
  On the other hand,
  \begin{align}
    \label{BoundF}
    f_2(L^m(v))+f_2(L^m(u))+f_2(L^m(vu))
    &\le 
    3f_2(L^m(vu)) + 2c_2 L^{>1}(vu)/L^m(vu)
    \nonumber
    \\
    &\le 
    3\kwad{f_2(\length{vu}) + c_2 \frac{L^{>1}(vu)}{L^m(vu)}}
   .
  \end{align}
  Inserting (\ref{BoundG}), (\ref{BoundF}), and (\ref{BoundL}) into
  (\ref{BoundI}) we obtain (\ref{IfBound}).
\end{proof*}

\section{Some properties of infinitary distributions}
\label{SecMath}

Infinitary distributions seem to be a~new interesting class of the
stochastic models for human narration.  The mathematics of excess
entropy is just being developed, cf.\
\citet{Debowski05} for an overview. Our
program is to bring together some advanced results of mathematics
(measure-theoretic probability theory, coding theory) and some
quantitative linguistic intuitions. We can give a~linguistic
interpretation to some mathematical theorems and a~formal language to
express some vague hypotheses about the obscure nature of
probabilistic language models.

We would like to mention four facts about infinitary distributions
which can be important for quantitative and computational linguistics
in the view of Hilberg's hypothesis.  These are:
\begin{enumerate}
\item There are infinitary distributions which are not deterministic
  stationary distributions. That is, total excess entropy $E=\infty$
  does not imply entropy rate $h=0$.
\item All stationary distributions which consist in a~random
  description of some infinite random object must be infinitary and
  nonergodic \citep[chapter 5]{Debowski05}. 

  Hence, we may suppose that $E=\infty$ holds for the stationary
  distribution of the language production because almost every human
  text refers  systematically to a~different and
  potentially infinite fictitious world.
\item For some infinitary distributions, value $\mathbf{P}(v)$ can be
  computed for every string $v$ by some finite procedure, cf.\
  \citet{Berthe94} and \citet{Gramss94}.
\item No infinitary distribution can be represented by a~finite-state
  hidden Markov model (HMM), cf.\ \citet{CrutchfieldFeldman01},
  \citet{Upper97}, \citet[section 2.8, data processing
  inequality]{CoverThomas91}.

  In spite of their inadequacy as the models of infinitary
  distributions, finite-state HMMs are the standard heuristic models
  of natural language engineering. It happens so only for the
  necessity of the effective search for the most probable hidden
  states.  Some well-known applications of HMMs are automatic speech
  recognizers \citep{Jelinek97} and trigram part-of-speech taggers
  \citep{ManningSchutze99,Debowski04}.  It was observed that the error
  rate of trigram taggers decreases as a~negative power of the size of
  the training data. When we increase the training data size ten
  times, the error rate diminishes only by half \citep{Megyesi01}.  In
  fact, such power-law decay of the error rate can be also some
  consequence of Hilberg's law \citep{BialekNemenmanTishby00}.
\end{enumerate}
The lack of space disallows us to exactly explain the terminology and
the reasons for the mathematical facts mentioned above.  We will try
to popularize some ideas of our thesis among the linguistic audience
in the next articles.

\end{document}